\documentclass[final]{cvpr}

\usepackage{times}
\usepackage{epsfig}
\usepackage{graphicx}
\usepackage{amsmath}
\usepackage{amssymb}
\usepackage{tabularx}
\usepackage{comment}

\usepackage[pagebackref=true,breaklinks=true,colorlinks,bookmarks=false]{hyperref}

\def\cvprPaperID{4060} 
\def\confYear{CVPR 2021}

\newcommand{\bai}[1]{\textcolor{black}{#1}}

\begin{document}

\title{Temporal-Channel Transformer for 3D Lidar-Based Video Object Detection in Autonomous Driving}


\author{
   {Zhenxun Yuan}${^{1}}$ \quad
   {Xiao Song}${^{2}}$ \quad
   {Lei Bai}${^{3}}$ \quad
   {Wengang Zhou}${^{4}}$ \quad
   {Zhe Wang}${^{2}}$ \quad
   {Wanli Ouyang}${^{1}}$\\
   ${^1}${The University of Sydney} \quad
   ${^2}${SenseTime Group Limited} \quad
   ${^3}${The University of New South Wales} \\
   ${^4}${The University of Science and Technology of China}\\
   {\tt\small zyua3118@uni.sydney.edu.au \quad songxiao@sensetime.com \quad baisanshi@gmail.com } \\
   {\tt\small zhwg@ustc.edu.cn \quad wangzhe@sensetime.com \quad wanli.ouyang@sydney.edu.au}
}

\maketitle

\begin{abstract}
The strong demand of autonomous driving in the industry has lead to strong interest in 3D object detection and resulted in many excellent 3D object detection algorithms. 
However, the vast majority of algorithms only model single-frame data, ignoring the temporal information of the sequence of data. In this work, we propose a new transformer, called Temporal-Channel Transformer, to model the spatial-temporal domain and channel domain relationships for video object detecting from Lidar data. 
As a special design of this transformer, the information encoded in the encoder is different from that in the decoder, \ie the encoder encodes temporal-channel information of multiple frames while the decoder decodes the spatial-channel information for the current frame in a voxel-wise manner.
Specifically, the temporal-channel encoder of the transformer is designed to encode the information of different channels and frames by utilizing the correlation among features from different channels and frames. On the other hand, the spatial decoder of the transformer will decode the information for each location of the current frame. 
Before conducting the object detection with detection head, the gate mechanism is deployed for re-calibrating the features of current frame, which filters out the object irrelevant information by repetitively refine the representation of target frame along with the up-sampling process. Experimental results show that we achieve the state-of-the-art performance in grid voxel-based 3D object detection on the nuScenes benchmark.
\end{abstract}

\section{Introduction}

\bai{3D object detection on Lidar-based point clouds plays an important role in robotic perception and autonomous driving applications. 
Although natural image and video based object detection has witnessed great improvements in recent years, recognizing and locating 3d objects from point clouds data remains challenging due to the irregular and uneven distribution of data points. To handle this problem, the main paradigm in the current 3D object detection approaches either utilize graph-based models directly on a raw point cloud frame \cite{2020PointRCNN,2019Fast,yang2019std} or transform 3D point cloud data into 2D pseudo image feature maps via Voxel Feature Encoding (VFE) methods and then conduct detection on a single frame \cite{DBLP:journals/corr/abs-1711-06396,article,2019PointPillars}. However, single-frame 3D object detection is still challenging (e.g., detect moving objects) due to the sparse nature of point clouds.
}

\begin{figure}[!t]
		\centering
		\includegraphics[width=1\linewidth]{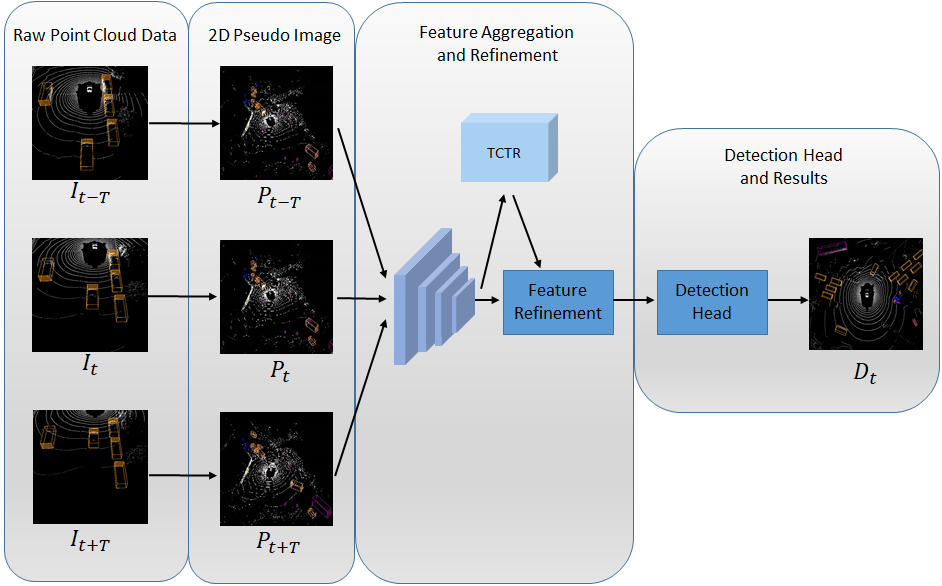}
		\caption{
			Framework overview. The raw data of continuous point cloud frames is converted into 2D pseudo-images. After feature extraction, the proposed TCTR module is adopted to generate representations containing temporal-channel information, followed by the feature refinement for the detection head.
		}
		\label{fig:Importance of temporal information}
		\vspace{-12pt}   
	\end{figure}
	

\bai{Utilizing contextual information from adjacent frames of point cloud video to enhance the target frame is a promising direction to solve the prominent data sparsity issue. 
In particular, early works \cite{hu2020get} in the area simply concatenate past point cloud frames with the current frame to make it more dense, which do not consider inter-frame correlations explicitly.
Following works further take spatio-temporal correlations among consecutive frames into account based on Recurrent Neural Network (RNN) or its variants, \ie. ConvGRU~\cite{convgru} and ConvLSTM~\cite{convlstm}.
All of these methods treat the information from all point cloud frames equally and integrate them together to get the final representation of the target frame, which is not precise and would introduce irrelevant information for two reasons. First, consecutive Lidar data are highly redundant due to the high sampling frequency (i.e., 20 point cloud frames per second with a 32-beam Lidar sensor \cite{2019nuScenes}). Second, Lidar also records massive information about the surrounding environment instead of the objects-of-interest. Thus, achieving a dense yet precise representation of the target frame based on the adjacent frames is a indispensable and critical problem for accurate Lidar-based 3D object detection. 
}

\bai{In this work, we study the 3D Lidar-based video object detection under the feature-based framework (as shown in Figure 1), which  projects the raw point clouds data to pseudo frames (i.e., 2D feature maps) considering the computation efficiency. Our work operates on several consecutive pseudo frames and attempts to achieve a dense and precise representation of the target frame by only integrating relevant information from the adjacent frames. }

\bai{To achieve this goal, we propose a Temporal-Channel Transformer module to reconstruct the target frame with both intra-frame and inter-frame relevant information in a fine-grained voxel-wised manner. The basic idea is using attention mechanism to find and integrate the useful information from correlated voxels in the target frame (intra-frame) and correlated frames in the input Lidar video (inter-frame) for each voxel of the target frame. Instead of deploying shallow attention layers, we adapt Transformer to the 3D Lidar video data analysis area by taking each channel of the compressed input frames as a Transformer encoder node and each voxel of the target frame as a Transformer decoder node, which enables to exploit diverse and complex spatial, temporal, and channel correlations among the whole input frames.
While the learned representation from Temporal-Channel Transformer is dense and only integrates information relevant with the target frame, it still contains object-irrelevant information as the sparse target frame and thus harms the detection performance. We solve this problem by combining the dense representation and sparse representation of the target frame together as the final representation with gating mechanism, which can control the information flow and consistently refine the final representation by removing object-irrelevant information.}

\bai{To summarize, we propose a new Lidar-based 3D video object detection method that achieves dense representation for the target frame with the help of adjacent frames. 
Our model has two main components, 1) a data enhancement module that integrates the most relevant information from the context frames to the target frame based on transformer in a voxel-wise manner, 2) a data refinement module that filters out object-irrelevant information based on the gate mechanism. 
Extensive experiments are performed on the large-scale nuScenes dataset \cite{2019nuScenes} to validate the effectiveness of our method. Our multi-frame model has achieved 7.4 mAP improvements and 5.1 mAP improvements over the single-frame baseline and state-of-the-art multi-frame 3D object detection method, respectively.}

\section{Related Work}

\subsection{Single-frame Lidar Object Detection}
3D Lidar-based single-frame object detection methods can be roughly divided into two groups: point-based \cite{2020PointRCNN,2019Fast,yang2019std},and grid-based \cite{DBLP:journals/corr/abs-1711-06396,article,2019PointPillars}. Point-based approaches are inspired by \cite{qi2017pointnet,qi2017pointnet++}, which capture features directly from the raw sparse point cloud data. Although the point-based approaches can lead to more accurate object detection, the high computational cost in finding the neighboring points makes them difficult to be used in real-time. The grid-based approaches \cite{DBLP:journals/corr/abs-1711-06396,article,2019PointPillars} convert the irregular point cloud data into a regular grid representation (e.g., voxels or pillars). Features are then extracted using 2D or 3D CNN. The overall framework of these grid-based works is generally the same, which consists of three parts. First, an Encoder for feature encoding, which projects point clouds into sparse pseudo-images under the bird's-eye view, such as the Voxel Feature Encoding layer (VFE) proposed by \cite{DBLP:journals/corr/abs-1711-06396} and the Pillar Feature Network(PFN) layers proposed in \cite{2019PointPillars}. 
\bai{Second, a feature representation module, which represents the pseudo-images with a well-designed network to capture complex spatial correlations in the point cloud frame. Third, a detection head with Region Proposal Network (RPN) for generating the final 3D bounding box.}

Considering the computational efficiency requirements of tasks using  3D object detection, we choose to transform the point cloud data into a representation of the grid pillars and then extract information from the grid pillars via 2D convolution. 

\subsection{Multi-frame Lidar Object Detection}
There is a growing number of works that use contextual information for 3D object detection. 
Hu et.al \cite{hu2020get} propose to fuse multiple frames into a single frame to expand the visibility region of the current frame. 
Ngiam et.al \cite{ngiam2019starnet} use the detection results of the previous frame as a prior knowledge to improve the detection at current frame. 
However, these works overlook the complex spatiotemporal dependencies among different frames, which have been demonstrated effective in natural 2D video object detection works. Recurrent neural networks and their variants (e.g., LSTM\cite{2012LSTM} and GatedRNN\cite{DBLP:journals/corr/ChungGCB14}) have achieved excellent results in sequence modeling and transduction problems, especially in the field of language modeling and machine translation. Applying recurrent structures to acquire temporal features has also driven the development of 3D video object detection \cite{2020Spatio,2017Object,8638831,DBLP:journals/corr/abs-1712-06317}. Yin et.al \cite{Yin_2020_CVPR} propose modules that fuse graph-based spatial coding features and combine them with spatial-temporal attention awareness modules to capture video coherence. Huang et.al \cite{huang2020lstm} uses 3D sparse convolutional neural networks to extract features from point cloud data and feed them into LSTM to produce hidden features and memorized features for continuous frame information transfer. 

Instead of using RNN-based temporal information processing, our work is the first (to our knowledge) investigating Transformer for 3D-Lidar-based video object detection. Transformer is adopted because it is known to perform better than RNN-based methods in sequential tasks like Natural Language Processing~\cite{vaswani2017attention}. 

\subsection{Transformer}
\bai{Transformer is proposed upon the attention mechanism in \cite{vaswani2017attention} to handle the sequence modeling problem by relating each node within a sequence to each other. It has achieved promising performance in several tasks, such as machine translation, document generation, and named entity recognition.} The core component of Transformer is the Multi-head scaled dot-product attention module, 
which achieves feature aggregation among all nodes. Compared to RNNs or CNN, multi-head attention has more powerful abilities in capturing global inter-dependencies among long range sequence.

Most recently, DETR \cite{carion2020endtoend} brings transformer into the computer vision field. This work integrates transformer into the 2D object detection framework and achieves higher accuracy than Faster RCNN \cite{fasterrcnn}. DETR extracts features from the input image using 2D CNN and then uses the Transformer for establishing the correlation among features from different locations for refining spatial features, which  are then used for predicting 2D detection results. 
Dosovitskiy et.al \cite{dosovitskiy2020image} also used the transformer model for the object classification task. They slice the input image grid into patches and flatten the patches through a linear projection matrix. Then, the transformer is used after adding the position embedding for feature aggregation. Finally, category recognition is performed through the classification head.

The encoder and decoder of transformers encode the same information in existing works for vision tasks. In contrast, the information captured and correlation built for encoder are different from that for decoder in our TCTR. Specifically, TCTR uses the encoder module for capturing temporal and channel information to model the relationship among features of different channels and frames, while the attention module in the decoder establishes the correlation between temporal-channel features of the encoder and spatial features for a better representation of the spatial features of the current frame. 

\begin{figure*}[!t]
	\centering
	\includegraphics[width=1\textwidth]{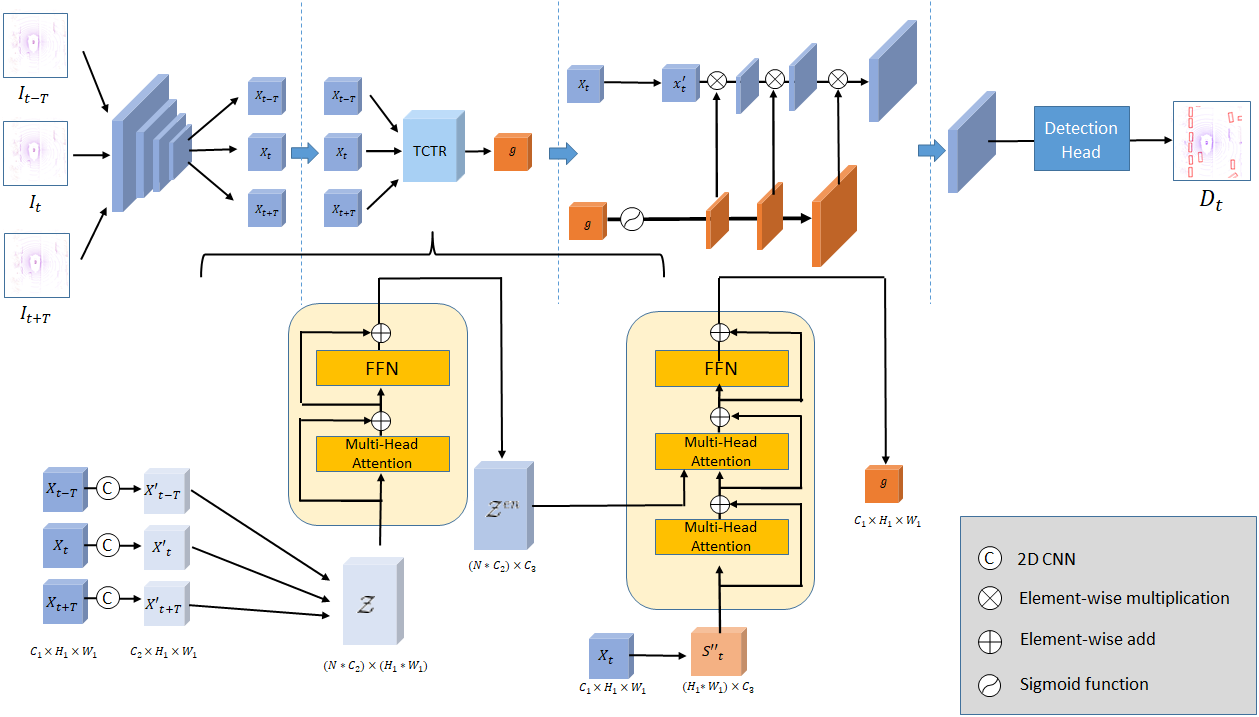}
	\caption{
		Network structure. We have a set of continuous point cloud frames $\mathcal{I}=\{I_{t-T},I_{t},I_{t+T}\}$ as inputs. Generate feature set $\mathcal{X}=\{X_{t-T},X_{t},X_{t+T}\}$ after backbone. $\mathcal{X}$  generates input for TC-encoder. $X_{t}$ from $\mathcal{X}$ generates $S^{''}_{t}$ as input to S-decoder. Our TCTR will generate a feature $g$ with temporal-channel information, then up-sample it synchronously with $X_{t}$ and refine $X_{t}$.
		} 
	\label{fig:training_diagram} 
\end{figure*}

\section{Method}
\bai{In this section, we present the details of our method. We first briefly introduce some preliminaries about 3D Lidar-Based video object detection and pre-operations (e.g., mapping 3D Lidar data to 2D pseudo image, backbone network for basic data representation) before our proposed network. 
Then, our Temporal-Channel Transformer is illustrated, which consists of a Temporal-Channel Encoder to explore channel-wise temporal correlations among all the input frames and a Spatial Decoder to enhance the target frame precisely with both intra-frame and inter-frame relevant information in a voxel-wise manner. 
In the next, the feature refinement module is presented, which combines the dense representation from Temporal-Channel Transformer with the original represention of the target frame through Gating mechanism to filter out object irrelevant information.
The detection head and training loss used in our work are discussed in the last.}

\subsection{Preliminary}
\bai{
The raw Lidar data generated at each time step is a sparse and irregular collection of point clouds frame, in which a point is represented as $\{x,y,z,r\}$, containing 3D space location information $\{x,y,z\}$ and reflection magnitude $r$. 
Denote the raw point clouds frame at time step $t$ as $I_{t}$. 
This work focuses on detecting objects-of-interest at $t$ with a set of continuous point cloud frames $\mathcal{I} =\{I_{t-T},...I_{t-1},I_{t},I_{t+1}...,I_{t+T}\}$.}

\bai{Instead of directly learning from raw point clouds data with point-based framework, we choose the voxel-based framework to ensure the computational efficiency.  
Specially, we first project the 3D space into $H_{0}\times W_{0}$ pillars under the bird view, and then extract features from the raw point clouds frame using PFN for each pillar. As a result, we get a 2D pseudo image for each point clouds frame and convert the frames $\mathcal{I}$ to 2D pseudo image sequence $\mathcal{P} =\{P_{t-T},...P_{t-1},P_{t},P_{t+1}...,P_{t+T}\}$, where $P \in \mathbb{R}^{C_{0}\times H_{0}\times W_{0}}$ and $T > 0$.}

\bai{Similar to natural videos, a sequence of consecutive point clouds frames also contains massive repetitive scenes, which make the pseudo images sequence $\mathcal{P}$ contain a lot of redundant information. 
Besides, there is a large difference between the scales of the objects-of-interest in the 3D space, such as pedestrians and construction vehicles, which makes it necessary to fuse the features with multi-scale receptive field. Thus, we will collect CNN features from different layers of backbone CNN with different resolutions. These CNN features are denoted by  $\mathcal{X} = \{X_{t-T},...X_{t-1},X_{t},X_{t+1}...,X_{t+T}\}, X\in \mathbb{R}^{C_1\times H_1\times W_1}$. 
As the backbone network is not our focus, we simply use two convolutional layers followed by four resblocks~\cite{he2015deep}, each resblock is followed by a max-pooling layer for downsampling.   
} 
 
\subsection{Temporal-channel Transformer}
\bai{Based on the CNN features $\mathcal{X}$, our target is to achieve a dense representation for the target frame by introducing relevant information from adjacent frames. 
Previous works on video object detection \cite{feichtenhofer2018detect,han2016seqnms,bertasius2018object,liu2018mobile} has demonstrated that exploring spatial and temporal correlations is useful for learning an accurate representation for the current frame and achieving better object detection performance. 
In our work, we further take channel-wise correlations into account considering that different channels could learn different patterns from the input, and explore spatial, temporal, and channel dependencies.}

\bai{We achieve this purpose by designing a novel Transformer-based data enhancement module, which consists of a Temporal-Channel Encoder and a Spatial Decoder. Specially, the Temporal-Channel Encoder is utilized to learn complex interactions among different frames based on the multi-head attention mechanism. Instead of viewing the whole frame as an entity, we regard each channel of a frame as a separate entity and feed all channels of the input sequence into the encoder, which can capture both intra-frame and inter-frame channel-wise dependencies. The Spatial Decoder is deployed to align the target frame with all channels of the transformed input sequence (i.e., output of encoder), which can achieve fine-grained feature aggregation with the most relevant information from the adjacent frames of the video in a pixel-wise manner.}

\subsubsection{Temporal-channel Encoder}

\bai{The TC-encoder adapts the structure designed in \cite{vaswani2017attention}, which has been shown to be effective in establishing relationships between nodes (e.g., words), into our task. 
Given the features $X_t$ of frame $t$,
we first obtain the transformed feature $X_t'$ of $C_2$ channels by transforming the features $X_t$ with a $1\times 1$ convolution operation. Then we flatten the transformed feature $X_t'$ along the spatial dimension for each channel to have stacked features $Z_t \in \mathbb{R}^{(C_2) \times (H_1\cdot W_1)}$. The features for all frames in $\mathcal{X}$ are obtained in the same way for $X_t$. In this way, we have the features for all frames in $\mathcal{X}$  and denote the features by
$\mathcal{Z}=\{Z_1, \ldots, Z_N\} \in \mathbb{R}^{(N \cdot C_2) \times (H_1\cdot W_1)}$ ($N=2*T +1$ denoting the input sequence length). 
In this stacked features, we treat each channel in $\mathcal{Z}$ as a node and flatten the spatial dimensions. Besides, a positional encoding is combined to $\mathcal{Z}$ to obtain the representation $\mathcal{Z}'$ with positional information in the same way as~\cite{vaswani2017attention}. Thus, the input to the multi-head attention block $\mathcal{Z}'$ can be regarded as a representation sequence containing $N \cdot C_2$ nodes. Each node represents a certain channel of one frame, and the feature of each node represent the spatial information among all $H_1 \cdot W_1$ voxels of the corresponding frame. We use the TC-encoder to utilize the  correlation among channel features within frames, as well as between frames.}
 

\bai{The TC-encoder consists of a stack of $M$ identical encoding blocks. Each encoding block contains a Multi-head self-attention layer and a Feature Forward Network (FFN) layer. 
The basic building block of multi-head attention is formulated as
}
\begin{equation}
Attention(Q,K,V)=softmax(\frac{QK^{T}}{\sqrt{d_{k}}})V.
\label{eq:Att}
\end{equation}
\bai{where $Q$ represents the queries, $K$ represents the keys, and $V$ represent the values, $d_k$ is a scalor for normalization. The match between query and key pairs, \ie $QK^{T}$, decides the attention score (i.e., importance) for values at the corresponding nodes. $softmax$ function is used to ensure the sum of importance values is 1. Instead of measuring the importance only once, multi-head attention attempts to capture more comprehensive interactions among different nodes and conduct multiple scaled dot-product attention in parallel. The results of all attention heads are combined together and once again projected with a linear transformation, leading to the final outputs as follows:}
\begin{eqnarray}
MultiHead(Q,K,V)=Concat(head_{1},..,head_{h}) \Theta ^{o}, \label{eq:MHAtt}\\
\textrm{where } head_{i}=Attention(Q_i, K_i, V_i), \label{eq:head} \\
Q_i = \mathcal{Z}\Theta_{i}^Q,\ K_i =  \mathcal{Z}\Theta_{i}^K,\  V_i = \mathcal{Z}\Theta_{i}^V. \label{eq:QKV}
\end{eqnarray}
\bai{
In our TC-encoder, $Q_i$, $K_i$, and $V_i$ are obtained from the transformations of $\mathcal{Z}$ with parameters $\Theta_{i}^{Q}$, $\Theta_{i}^{K}$, and $\Theta_{i}^{V}$, respectively. Thus, multi-head attention can capture both inter-channel and inter-frame correlations. 
$\Theta ^{o}$ are parameters for the output projection.}


\subsubsection{Spatial Decoder}

\bai{The spatial decoder module focuses on enhancing the features $X_t$ of the current frame with both intra-frame and inter-frame relevant information.
}

For the intra-frame spatial correlation, we first apply a $1 \times 1$ convolution operation to $X_t$ and obtain transformed features $\hat{S}_t \in \mathcal{R}^{C_3 \times H_1 \times W_1}$. Then, $\hat{S}_t$ is reshaped to a matrix of size $\mathcal{R}^{(H_1 \cdot W_1) \times C_3}$ and combined with the positional embedding.
Here, we directly use the position encoding method performed in \cite{parmar2018image,bello2020attention} as it is widely adopted in existing works. 
Denote $S_{t}'$ as the feature concatenating reshaped feature $S_t$ and position information. 
Similarity to the encoder, multi-head attention is also applied to the decoder. 
There are two types of attentions in the decoder. 

As the first type, self-attention is used. We set both queries, keys, and values as the transformation of $S_{t}'$ and calculate the correlation among different nodes. 
After this multi-head attention, each voxel in $X_t$ is enhanced with the information from the most relevant voxel in the same frame and $S_t$ now becomes $S_t'' \in \mathcal{R}^{(H_1 * W_1) \times C_3}$.

\bai{After the self-attention operation, $S_t''$ is further fed to the second type of multi-head attention block to capture inter-frame spatial correlations and enhance  $S_t''$ with inter-frame information. 
Denote the output of the encoder by $\mathcal{Z}^{en}$.
Specifically, the formulation of Eq.~(\ref{eq:QKV}) for $Q_i, K_i$, and $V_i$ in the encoder is implemented in the second type of attention in the decoder as follows:
\begin{equation}
Q^{d2}_i = S_t''\Theta_{i}^{d2, Q},\ K^{d2}_i =  \mathcal{Z}^{en}\Theta_{i}^{d2, K},\  V^{d2}_i = \mathcal{Z}^{en}\Theta_{i}^{de, V}. \label{eq:decatt}
\end{equation}
In Eq.~(\ref{eq:decatt}), the query $Q^{d2}_i$ is from the features $S_t''$ of current frame and the key $K^{d2}_i$ from the output of the encoder features $\mathcal{Z}^{en}$. 
To achieve more fine-grained spatial interactions, we attempt to measures the relationships between each voxel of current frame $t$ with all frames in $\mathcal{X}$ in the channel level instead of the frame level, which is achieved by aligning $S_t''$ with the output of the Temporal-Channel Encoder module $\mathcal{Z}^{en} \in \mathcal{R}^{(N\cdot C_2) \times C_3}$. As shown in Figure \ref{fig:training_diagram}, we set the transformation of $S_t''$ as the query $Q$, and set the projection of $\mathcal{Z}^{en}$ as the keys and values of the multi-head attention block. Through the attention mechanism, all channels in the the input sequence are matched with each voxel in $S_t''$ and contribute to the new representation of the voxel according to their correlations. Thus, the final output of the Spatial decoder is the enhanced $X_t$ with both inter-frame and intra-frame information considering spatial correlations, temporal correlations, and channel correlations. 
}

\begin{figure*}[!t]
	\centering
	\includegraphics[width=1.00\textwidth]{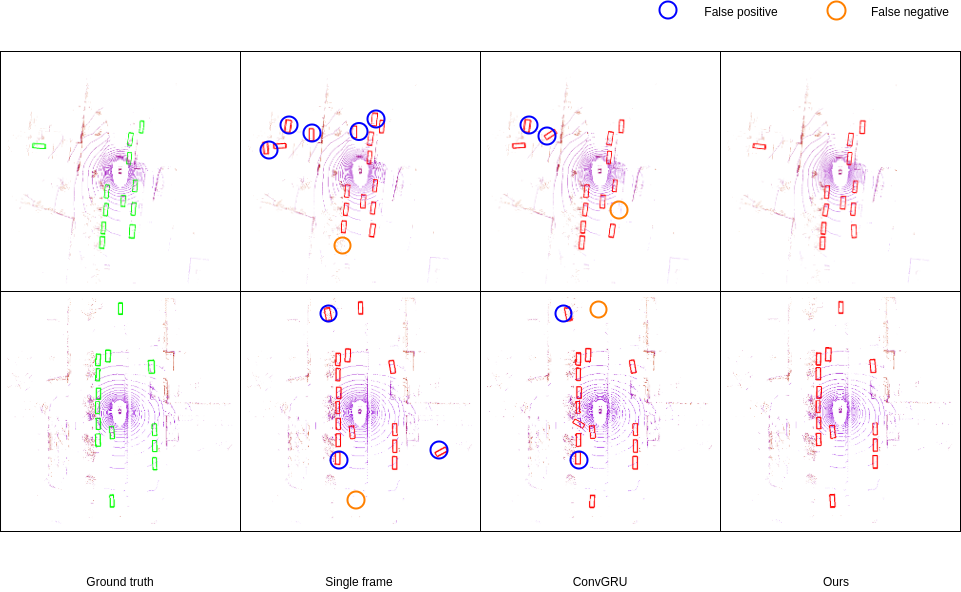}
	\caption{
	    Comparison between our method, the single-frame baseline, and the ConvGRU-based method. The green box represents the ground truth. The red box indicates the test results. From left to right, the first, second, third, and forth columns are the ground truth, the detection results for the single frame baseline, the ConvGRU, and our proposed TCTR, respectively.
		} 
	\label{fig:result_visualization} 
\end{figure*}

\subsection{Feature Refinement Module}
\bai{The output of the Temporal-Channel Transformer module $g \in \mathbb{R}^{C_1\times H_1\times W_1}$ is a more dense representation than $X_t$ and contains spatial, temporal, and channel dependencies among the input frames. 
Instead of feeding $g$ into the object detection head directly, we combine $X_t$ with $g$ to integrate more useful information. 
It could also lead to more accurate features of $X_t$ and naturally benefit $g$. Other than simply concatenating or adding $g$ with $X_t$, we use the gate mechanism to combine them together, which is widely used in sequence modeling (e.g., RNN) and has the capability to control the information flow and filter out the information irrelevant to object~\cite{dauphin2017language}. Formally, the gate mechanism based refinement is defined as:}
\begin{equation}
    F_{t}=X_{t} \otimes \sigma(g)
    \label{eq:gate}
\end{equation}
\bai{where $\sigma$ is the sigmoid activation function, $\otimes$ is the element-wise multiplication.}

\bai{We up-sample the re-calibrated representation $F_{t}$ to the same resolution as $P_t$ for conducting the detection with a multi-layer CNN structure. Along with the up-sampling process, we also up-sample the dense representation $g$ and consistently combine the up-sampled $g$ with the up-samped $F_t$ to keep refining the output and ensure the feature space consistency.}


\subsection{Detection Head and Training Loss}

\bai{We use the same detection head and loss function as in PointPillars \cite{2019PointPillars}. Specially, 2D IoU is used to select the prior-box for ground truth, regardless of the height dimension of the object to be measured. Focal loss is used for object classification:} 

\begin{equation}
\mathcal{L}_{cls}(P_{t})=-(1-P_{t})^{\gamma }log(P_{t})
\label{fl}
\end{equation}

In addition to generating the object category, we also need to obtain the 3D position $(x,y,z)$ of the object, as well as the box's length ($l$), width ($w$), height ($w$), and orientation angle ($o$).
We use the Smooth L1 loss to get the location, which is most commonly used in object detection tasks.
\begin{equation}
\mathcal{L}_{loc}=\sum_{b\in (x,y,z,l,w,h,o)} SmoothL1(\Delta b)
\label{smoothl1}
\end{equation}

Although we have regressed the values of the orientation angles, we also need to make positive and negative judgments about the direction of the orientation angles to further correct the orientation angles. We use softmax classification loss $ \mathcal {L}_{dir}$ to optimize orientation angle judgments.

Thus, the total loss is:
\begin{equation}
\mathcal{L}=\frac{1}{n}(\beta_{cls}\mathcal{L}_{cls}+\beta_{loc}\mathcal{L}_{loc}+\beta_{dir}\mathcal{L}_{dir} )
\label{smoothl1}
\end{equation}
\bai{where $n$ stands for number of positive anchors, and $\beta_{cls}$, $\beta_{loc}$, and $\beta_{dir}$ represent the weight of the three loss terms, respectively.}

\begin{table*}
    \centering
    \caption{We validate our method on the nuScenes dataset \cite{2019nuScenes}. T.C, Moto and Cons present traffic cone, motorcycle and construction vehicle. The results are divided into  single-frame approach and  multi-frame approach. The results show that we have achieved state of the art in both grid voxel-based single-frame and multi-frame methods.
    }
    \resizebox{\textwidth}{!}{
    \begin{tabular}{c|c| c|c|c|c|c|c|c|c|c|c|c} 
    \hline\hline
      Type&Method  &  Car   & Ped.  &   Bus  &  Barrier   &  T.C  &  Truck  &  Trailer   & Moto.   &  Cons   &  Bicycle & Mean  \\ \hline
        
       &MAIR\cite{simonelli2019disentangling} & 47.8 & 37.0 & 18.8 & 51.1 & 48.7 & 22.0 & 17.6 & 29.0 & 7.4 & 24.5 & 30.4  \\ 
       &PointPillars\cite{2019PointPillars} & 68.4 & 59.7 & 28.2 & 38.9 & 30.8 & 23.0 & 23.4 & 27.4 & 4.1 & 1.1 & 30.5 \\ 
       Single-frame&SARPNET\cite{YE202053} &59.9&69.4&19.4& 38.3& 44.6& 18.7&18.0&29.8& 11.6&14.2&32.4\\ 
       &WYSIWYG\cite{hu2020get} &79.1&65.0&46.6& 34.7& 28.8& 30.4&40.1&18.2&7.1&0.1&35.0\\ 
       &SSN\cite{zhu2020ssn} &80.7&72.3&39.9& 56.3& 54.2&37.5&43.9&43.7&14.6&20.1&46.3\\ 
       &PointPainting\cite{vora2020pointpainting} &77.9&73.3& 36.1& 60.2&62.4&35.8&37.3&41.5&15.8&24.1&46.4\\ \hline
      multi-frame&3DVID\cite{Yin_2020_CVPR} &79.7&76.5&47.1 &48.8& 58.8& 33.6&43.0&40.7& 18.1&7.9&45.4\\ 
       &Ours &83.2&74.9&63.7&53.8&52.5&51.5&33.0&54.0&15.6&22.6& \textbf{50.5}\\ 
      \hline\hline
    \end{tabular}}
    \label{tab:overall}
\end{table*}

\section{Experiment}

\subsection{Datasets}
\bai{Most previous single-frame 3D object detection methods conduct experiments on the KITTI dataset \cite{2013Vision}. However, KITTI dataset does not have continuous frames, which make it infeasible for evaluating 3D video object detection methods. We evaluate our model on the nuscnese dataset \cite{2019nuScenes}, which has a richer set of 700 scenes and a larger amount of data. The nuscnese dataset includes 700 scenes for training and 150 scenes for testing. The nuscnese dataset contains 20 frames for each second and is annotated every 0.5 second. We denote the annotated frames as key-frames and others as sweeps.}


\subsection{Implementation Details}
The $x$, $y$, and $z$ coordinates of the point clouds are in the range of $-51.2$ to $51.2$, $-51.2$ to $51.2$, and $-5.0$ to $3.0$, respectively.    
Each point cloud is divided into $512\times512$ pillars with each has a size of $0.2\times0.2\times8$.
As we have mentioned above, FPN is first used to derive features for each pillar.
The output size of the feature maps of FPN is $64\times512\times512$.
These feature maps then pass through the backbone to generate the feature maps with size of $16\times16\times256$, which finally go through the TCTR and feature refinement modules.
The anchor size we set for the bounding box for regression is the mean of the ground truth. 
The weights $\beta_{cls}$, $\beta_{loc}$, and $\beta_{dir}$ in our proposed loss function are set to $1$, $0.25$, and $0.2$, respectively. 
We use the Adam optimizer and one-cycle training strategy with a learning rate of 0.001. 
We train 40 epochs with 16 GPUs.
The batch size is set to 2.
The number of frames $N$ is set to $3$ to achieve a good balance between the performance and complexity.

Single frame 3D object detection algorithms usually add ground truth of one frame to another frame for training data augmentation \cite{zhu2019classbalanced}. 
Our proposed method, however, considers the temporal correlation of objects in neighboring frames.
Therefore, we cannot use this kind of training data augmentation.
Instead, we use the random flipping along the x, y-axis, and rotating along the z-axis in the range of $-0.3925$ to $0.3925$. 
We also apply a random scale ranging from $0.95$ to $1.05$ for data augmentation. 

\subsection{Overall Performance}
\bai{We present the quantitative results of our model and other state-of-the-art methods on the nuScenes dataset in Table \ref{tab:overall}. Among the comparison methods, PointPillars \cite{2019PointPillars}, SARPNET \cite{YE202053}, and WYSIWYG \cite{hu2020get} are grid voxel-based single frame 3D object detection methods. PointPainting \cite{vora2020pointpainting} is based on the PointPillars framework and further fuse the Lidar data with natural image data to provide richer information. 3DVID \cite{Yin_2020_CVPR} is the state-of-the-art 3D video object detection network on the nuScenes dataset, which combines KNN graph and ConvGRU to capture the spatiotemporal correlations. SSN\cite{zhu2020ssn} takes advantage of the shape features of the point cloud to propose a way to encode the shape of the point cloud to improve the object classification.
}

\bai{Our model achieves the best performance among all the comparison methods with a large advantage. In particular, we outperform the 3DVID model and original PointPillars approach by 11.2\% (5.1 in mAP) and 65.6\% (20.0 in mAP), respectively. Although our method only utilizes the point clouds data in the nuScenes dataset, we still achieves better performance than PointPainting, which used both Lidar and natural image data source.}

\subsection{Ablation Studies}
\subsubsection{Ablation Study A}
\bai{we first conduct an overall ablation study about our model. As shown in Table \ref{ablation_a}, the baseline of our method only contains the backbone network and multiple CNN up-sampling layers (directly sets $F_t$ as $X_t$ as Equation \ref{eq:gate}), which is a single frame 3D object detection method. The "baseline+concat" extend "baseline" to video object detection simply by concatenating $\mathcal{X}$ as $F_t$. The "baseline+TCTR" variant only uses Temporal-Channel Transformer to explore the spatial, temporal, and channel dependencies among the input frames and sets $F_t$ as $g$ for detection. We can observe that "baseline+concat" performs better than baseline, which demonstrates the importance of exploring adjacent frames for 3D object detection. Besides, integrating adjacent frames with our TCTR module is better than simply concatenating the inputs together, which shows the necessity of exploring the complex correlations among the sequence. Our feature refinement module can further re-calibrate the learned representation and help improvement the object detection performance.
} 

\begin{table}[]
\centering
\caption{Ablation Study A: overall ablation study on the framework.}
\label{ablation_a}
\begin{tabular}{c|c|c}
\hline
Method     & Frames & mAP   \\ \hline
baseline         & 1      & 43.15 \\
baseline+concat  & 3      & 45.57 \\
baseline+TCTR     & 3      & 48.43 \\ \hline
baseline+TCTR+FRM (ours) & 3      & \textbf{50.47} \\ \hline
\end{tabular}
\end{table}

\subsubsection{Ablation Study B}
\bai{We then conduct a deeper study regarding the effectiveness of our Temporal-Channel Transformer by replacing it with other networks or variants. Specially, we consider the following four comparison methods: 
1) ConvLSTM as used in \cite{convlstm}, 
2) ConvGRU as used in \cite{convgru}, 
3) T-encoder which does not consider the channel-wise correlations by taking the whole representation of each frame as a node for encoding (e.g.,, $X_t'$ instead of a channel of $X_t'$),
4) C-endoer which does not consider the temporal correlation and only take different channels of $X_t'$ as nodes for encoding.
Expect the Temporal-Channel Transformer part is replaced, all other parts of our model are keep unchanged. 
As can be observed from Table \ref{ablation_b}, our method achieves the best performance among all the compared variants, which demonstrate the superior of our Temporal-Channel Transformer in capturing complex dependencies in the input  images. Besides, our Temporal-Channel encoder achieves better performance than T-encoder, revealing the importance of exploring channel-wise correlations.}

\begin{table}[]
\centering
\caption{Ablation Study B: ablation study on the Temporal-Channel Transformer.}
\label{ablation_b}
\begin{tabular}{c|c}
\hline
Method     & mAP   \\ \hline
ConvLSTM   & 46.92 \\
ConvGRU    & 47.41 \\
T-encoder  & 48.71 \\
C-encoder  & 47.43 \\ \hline
TC-encoder (ours) & \textbf{50.47} \\ \hline
\end{tabular}
\end{table}

\subsubsection{Ablation Study C}
\bai{In this part, we study the influence of different methods for combining $X_t$ and $g$, as discussed in Section 3.3. We compare our gate mechanism based feature refinement module with concatenation and adding based fusion strategies. The result without combining $X_t$ is also presented for comparison. 
As shown in Table \ref{ablation_c}, the detection accuracy can be improved no matter combining the information from $X_t$ with concatenation or adding, which spots the importance of current frame in object detection.
Besides, our feature refinement module uses gating mechanism to select the detection relevant information and further achieves better performance, which demonstrates the effectiveness of our design and highlight the importance of paying more attention on object-relevant features.
}
	
\begin{table}[h]
    \centering
    \caption{Ablation Study C: ablation study on the feature refinement module, $||$ denotes the concatenate operation.}
        \begin{tabular}{c | c}
	        \hline
	        Method & mAP  \\
	        \hline
	        $F_t = X_t$ & 48.43 \\
	        $F_t = X_t || g$ & 49.21 \\
	        $F_t = X_t + g$ & 49.37 \\
	        $F_t = X_t \otimes \sigma(g)$ (ours) & \textbf{50.47} \\
	        \hline
        \end{tabular}
    \label{ablation_c}
    \end{table}

\subsubsection{Ablation Study D}
\bai{How the number of input frames influences the object detection performance is also of our interest. In this study, we evaluate our model's performance with different numbers of input frames. As mention in Section 4.1, nuscnese dataset contains both keysframes and sweeps. We only use the keyframes without any sweeps as input for this study due to the GPU memory and training time constraints. The results are presented in Table \ref{ablation_d}. It can be observed that the detection accuracy consistently improves with the input length, revealing the importance of exploiting and aggregating the information from more frames to make the target frame more dense. }
	
\begin{table}[h]
    \centering
    \caption{Ablation Study D: ablation studies regarding the relationship between frames number and performance.}
        \begin{tabular}{c | c}
	        \hline
	        Input Lengths & mAP  \\
	        \hline
	        1T & 40.62 \\
	        3T & 42.05 \\
	        5T & 43.76 \\
	        7T & 44.68 \\
	        \hline
        \end{tabular}
    \label{ablation_d}
    \end{table}

\section{Conclusions}

\bai{In this work, we study the 3D Lidar-based video object detection problem and propose a novel deep learning method for enhancing the target point clouds frame with adjacent frames. A new transformer, named Temporal-Channel Transformer module, is designed to fully explore the spatial, temporal, and channel correlations among different frames and enhance the target frame based on these learned correlations in a voxel-wise manner. Besides, we also design a feature refinement module to re-calibrate the learned dense representations, which helps filter out the object irrelevant information. We conduct experiments on the large-scale nuScnese dataset and compare our method with several strong baselines, which demonstrate our model achieves state-of-the-art performance. Extensive ablation studies are also conducted and demonstrate the effectiveness of our design.}

\bibliographystyle{ieee_fullname}
\bibliography{egbib}

\end{document}